# Improving fraud prediction with incremental data balancing technique for massive data streams

Rafiq Ahmed Mohammed
Murdoch University
Perth, Australia
Rafiq.mohammed@murdoch.edu.au

Kok-Wai Wong
Murdoch University
Perth, Australia
K.wong@murdoch.edu.au

Mohd Fairuz Shiratuddin
Murdoch University
Perth, Australia
F.shiratuddin@murdoch.edu.au

Xuequn Wang
Murdoch University
Perth, Australia
A.wang@murdoch.edu.au

*Abstract*—The performance of classification algorithms with a massive and highly imbalanced data stream depends upon efficient balancing strategy. Some techniques of balancing strategy have been applied in the past with Batch data to resolve the class imbalance problem. This paper proposes a new incremental data balancing framework which can work with massive imbalanced data streams. In this paper, we choose Racing Algorithm as an automated data balancing technique which optimizes the balancing techniques. We applied Random Forest classification algorithm which can deal with the massive data stream. We investigated the suitability of Racing Algorithm and Random Forest in the proposed framework. Applying new technique in the proposed framework on the European Credit Card dataset, provided better results than the Batch mode. The proposed framework is more scalable to handle online massive data streams.

## I. INTRODUCTION

Within the machine learning advancement, automated balancing strategy for highly imbalanced data has emerged as a research focus area. The dataset is deemed to be imbalanced if one of its classes exhibits a huge dominance over the other class [1], this issue is common for the fraud detection problem. The machine learning classification techniques used in the process of credit card fraud detection struggle with massive imbalanced datasets, due to the large computation required to predict fraud [2]. Popular machine learning toolkits like Weka [3], R [4] and scikit-learn in Python [2] has the pre-processing techniques to resolve the problem of class imbalance. In addition, the researchers [5] put an emphasis on using some kind of automatically searching techniques to optimise from the data and propose a good balancing technique such that it can maximize the classifier performance. The process of adapting to select an appropriate balancing technique would resolve the overarching research question on "how to push the boundaries of fraud prediction on the underrepresented minority class?" [6]. However, the literature suggests that most of the automated balancing solutions are available for the Batch data [4], i.e. when all data are available and balanced them as Batch.

In real-time, credit-card transactions follow non-stationary [7] and a sequential pattern [8], which evolves over time because of seasonality and new attack strategies [9]. In addition, when the data streams become highly imbalanced the frequency of fraud class becomes less and less frequent resulting with an enormous imbalance ratio [10]. The prediction models in order to learn with imbalanced data streams require balancing techniques which can dynamically adapt to make the data balanced with ratio of 1:1 between normal and fraudulent class [11]. This process of effective balancing also termed as re-balancing technique will make the classification technique to be adaptive for online streaming data [12] for better fraud prediction. For these reasons, there is a need for new data balancing framework which can handle massive data streams. The performance of classification algorithms depends upon whether they are suitable to handle massive data, and can work with data streaming environment, suitable classification technique is proposed.

Ensemble learners are considered effective for data stream classification problems [13]. Online Random Forest (ORF), an ensemble learner for data stream classification, is insensitive to feature scaling [14] and is considered very effective with high accuracy [15]. The Mondrian forest, one variant of ORF, can be used for training the dataset in an incremental fashion and has excelled in online learning with large Batch problems [16]. In addition, Tree ensembles are very popular for incremental online learning, due to their high accuracy, simplicity and parallelization capability [15]. Moreover, a combination of boosting algorithms and Synthetic Minority Over-Sampling Technique (SMOTE) can apply different weights to the data instances dynamically [17], and can successfully address the problem of the class imbalance problem [6]. Also, Random Forest (RF) with Random Under Sampling (RUS) balancing technique is proven to be scalable and capable of Fraud Detection (FD) with highly imbalanced massive datasets [2]. In addition, RF with SMOTE is considered as a best technique for the European Credit Card (ECC) dataset [5]. On the other hand, Racing Algorithm (RA), an automated data balancing technique is available in R toolkit [4], which optimizes the process of finding the right re-balancing technique for any dataset [5]. In this work, we implemented *"Racing Algorithm"* [18, 19], which runs all the available balancing techniques in parallel as a pre-selection phase to select the optimal re-balancing technique for ECC dataset and to achieve highest performance metrics for RF classification technique.

In this paper, a new framework for incremental data balancing for fraud detection classification problem that can work with data streaming environment is proposed. The proposed framework is named as Piece-Wise Incremental Data Balancing (PWIDB) framework. In order to demonstrate the proposed framework, RA is used as the data balancing technique for the data streaming environment. RF, which is a popular classification technique used for massive data streams will also be used in the proposed framework. Therefore, in this paper we investigated the suitability of RA and RF in the proposed framework, and we argue that incremental data balancing framework is suitable for incremental learning and will generate efficient models for fraud prediction.

In the process of investigating whether RA works in the proposed framework, the results of the RA used for Batch data, will also be used for comparison. The strategy in the RA is to adapt to the given dataset, performance metric used and the classification algorithm [5] to select the best balancing method. The researcher [19] re-iterated that a good balancing technique will solve the class imbalance problem. The experimental results as discussed in section IV-E-1, have shown that RA is effective to be used as an automated re-balancing technique for the Batch mode of a learning environment.

Based on our success of RA with the Batch mode learning environment, we investigated whether RA can also be used for data streaming environment in the proposed framework. This investigation process seeks to extend a perceived gap by exploring the impact of automated re-balancing methods on the classification results for incremental learning. To the best of our knowledge, only the RA has implemented the automated balancing strategy in batch data, and has not been implemented in any incremental streaming Batch nor for the incremental online learning [20]. For this reason, we organized the remaining of this paper as follows: in Section II there is some explanation about the credit card fraud detection problem formulation, to be addressed in a real-time Fraud Detection System (FDS), and we explain our proposed framework. In Section III we give some explanation about adaptive and online balancing techniques. In Section IV we explain our new fraud detection technique. This section is further subdivided into five subsections. In Subsection IV-A, we summarise the dataset used for the experiments. In Subsection IV-B, we discuss different data re-balancing techniques to tackle the problem of class imbalance. In Subsection IV-C, we discuss the experimental setup. Subsection IV-D describes the RA technique. Subsection IV-E shows the experimental procedure for automated balancing strategy methodology and the RA implementation. For simplification, the experimental procedures are further subdivided into three subsections. Subsection IV-E-1 shows RA implementation in Batch mode and presents a comparative experiment with RF without re-balanced data. The experimental results as discussed in Subsection IV-E-1, have shown that RA is capable as an automated re-balancing technique for the Batch mode of the learning environment. Subsection IV-E-2 shows RA implementation in data streaming mode and presents a comparative experiment with RF without re-balanced data. The results will be discussed in subsection IV-E-2. The subsection IV-E-3 shows the implementation of the new incremental balancing technique using RA in data streaming mode using the proposed PWIDB framework, and presents a comparative discussion about the results achieved by our new technique using the proposed framework. Finally, we conclude in section V.

## II. Problem Formulation

The basic idea of our proposed Piece-Wise Incremental Data Balancing (PWIDB) framework is to create a new environment to fit incremental re-balancing technique with a compatible classifier technique to work in a real-time streaming FDS. A simple learning scenario of credit card fraud detection classification problem, to be addressed in a real-time FDS is formulated by considering the following steps: the classifier technique '$C$' is used to process feature vectors $d_1, d_2,...,d_j$ in a chunk of transactions $D_j$ received at time $T_j$, and $H_j$ is the hypothesis of model generated for the classifier '$C$'. As depicted in Fig. 1, the streaming Batch learning models $M_j, M_{j+1},...,$ are generated for streams of data chunks $D_j, D_{j+1},...,$. We denote '$b$' as automated data balancing technique, '$p$' as a performance metric (F1, AUC), and '$C$' as the classifier. The variable '$b$' is only used when RA or any data balancing optimization technique is used. If we just use RF without data re-balancing then '$b$' is assigned with 'unbal', along with $S_j$ is assigned with 'no re-sampling'. The learning models adapt to the Imbalance Ratio (IR) rates $I_j, I_{j+1},...,$ along with re-balanced data value '$b$' within the data streams $D_j, D_{j+1},...,$ and subsequently, the hypotheses $H_j, H_{j+1},...,$ is generated.

In stream Batch, a classifier model $M_j$ work with $W_j$ window of the dataset. A new stream Batch learning model $M_{j+1}$ is created with the arrival of a new stream of data chunk $D_{j+1}$, and for the new active window which will be:

$$W_{j+1} = D_{j+1} \qquad (1)$$

A new hypothesis $H_{j+1}$ is generated when the IR ratio $I_{j+1}$ is changed with the transactions in the new active window $W_{j+1}$ as compared to $I_j$. The derived hypothesis $H_{j+1}$ is also dependent on the data balancing technique '$b_{j+1}$' selected for active window of dataset $W_{j+1}$ and based on the performance metric '$p$' and the classifier '$C$'.

$$H_j = \{d_j \in D_t \; p.c.b \; I_j(D_j) \leq m\} \qquad (2)$$

We define the hypothesis $H_j$, for which the RA during the process of incremental learning, automatically select whether re-balancing of current active window $W_{j+1}$ is required, and the re-balancing decision is made to enhance the prediction results. This decision of re-balancing is derived based on the IR of $I_{j+1}$ of $W_{j+1}$ which can be greater than $I_j$, and the best re-balancing technique '$b$' is selected based on its statistical performance.

If the derived hypothesis $H_j$ in equation (2) ascertain that for e.g. the SMOTE re-balancing technique '$b_j$' selected for active window of dataset $W_j$. So, now the classifier model $M_j$ working with $W_j$ window of the dataset is updated incrementally with the arrival of new data chunk $D_{j+1}$, along with "re-balanced" data chunk $S_j$ from window $W_j$ and the new active window instead of (1) will be:

$$W_{j+1} = D_{j+1} + S_j \qquad (3)$$

The main characteristics of an incremental data balancing technique are:
- The incremental data balancing technique to derive the best approximation.
- With streaming Batch learning models as learning improves the technique derives new hypothesis $H_j$ at time $j$.
- As learning improves the incremental re-balancing technique helps to derive new hypothesis $H_{j+1}$ at time $j+1$.

This new hypothesis works best than normal streaming Batch learning hypothesis and its classification models.

The proposed PWIDB framework is explained here with a working example. As depicted in Fig. 1, and as tabled in VI, the technique when working with the data chunk window $W_j \leftarrow D_j$, the RA with RF classifier model $M_j$ derives hypothesis $H_j$ as {$b_j$=SMOTE, $S_j$=Over sampling value(1575), '$p$'=AUC, '$C$'=RF, $I_j$=1.9, $W_{jTr} \leftarrow D_{jTr}$}. This hypothesis signifies that the Synthetic Minority Over Sampling Technique (SMOTE) has been selected as the balancing technique $b_j$ to work with the classifier model $M_j$, and for the dataset $W_j$. The derived IR value of 1.9 for $I_j$ after applying the balancing technique $b_j$, for this hypothesis $H_j$ will generate the best prediction results. In addition, after applying the balancing technique $b_j$, the data is re-balanced and the derived re-balanced data of $W_j$ has been assigned to the re-balanced variable $S_j$. Since, this re-balanced data value of 1575 for $S_j$ has contributed to improve the AUC value to more than 0.99, the re-balanced data $S_j$ value from $W_j$ will be used to increment the next window $W_{j+1}$ value.

Similarly, the PWIDB framework when working with the data chunk window $W_{j+1}$, the training and validation data is derived as $W_{j+1Train} \leftarrow D_{j+1Train}+S_j$. i.e. 46575←45000+1575. The RA with RF model $M_{j+1}$ derives the hypothesis $H_{j+1}$ as {$b_{j+1}$=SMOTE, $S_j$=Over sampling value - 1575, '$p$'=AUC, '$C$'=RF, $I_j$=4.3, $W_{j+1Tr} \leftarrow D_{j+1Tr}+S_j$}. This hypothesis signifies that SMOTE has been selected as the balancing technique to work with the classifier model $M_{j+1}$, and for the dataset $W_{j+1Tr}$, and for the IR $I_{j+1}$ is 4.3. This implies that hypothesis $H_{j+1}$, will generate the best prediction results for the window of dataset $W_{j+1}$.

## III. ADAPTIVE AND ONLINE BALANCING TECHNIQUES

Incremental learning techniques are used to generate hypotheses with the training samples, whereby the learner can use the hypothesis at any time to generate and validate the best approximation answer to the query [21]. In similar fashion, the proposed technique of PWIDB generates and validate the hypothesis for each incremental window to generate the best prediction results. Since knowledge evolves over time, these techniques need to adapt to improve intelligence [22] and to generalize new hypotheses. This ubiquitous learning of incremental tasks by the techniques [21] will enhance their ability to learn continuously and improve performance overtime [23]. The technique of Batching the data reduces communication overhead with data streams [24]. Correspondingly, the assumption made in this paper is that similar to communication problem which relies on a stream of information, a good balancing technique should adapt and learn incrementally to resolve class imbalance problem for massive data streaming situation. Researchers show that the techniques of up-sampling or down-sampling do not resolve the problem of imbalanced datasets [25]. However, with the implementation of incremental setting in the process of active learning, it is able to adjust a re-balancing strategy dynamically [25]. In the process of adaptive learning strategy to resolve class imbalance problem, we analysed the techniques available to implement the automated balancing strategy. The researchers [26] proposed pre-processing techniques based on SMOTE to deal with imbalanced data streams, and these techniques use minority class data to update the base learner incrementally [6]. However, the literature is limited to incremental algorithms to manage finite datasets [26]. Data streams are unbounded and the ordered sequence of instances that arrive over time [27]. Moreover, the instances are not given beforehand but become available sequentially or in the form of data chunks (blocks) as the stream progresses [26]. Incremental learning builds up the knowledge from the online streaming data to make informed future predictions [28]. Some researchers suggested that during the process of incremental learning, machine learning algorithms should adapt to learn from infinite historical data [29]. Other researchers argue that the learner has access to partial historical data [30], and this technique of preserving previously acquired knowledge does not suffer from catastrophic forgetting [15]. However, this process limits its capability to access historical data. The work in this paper utilised a methodology to retain and access the historical data for the incremental learning environment.

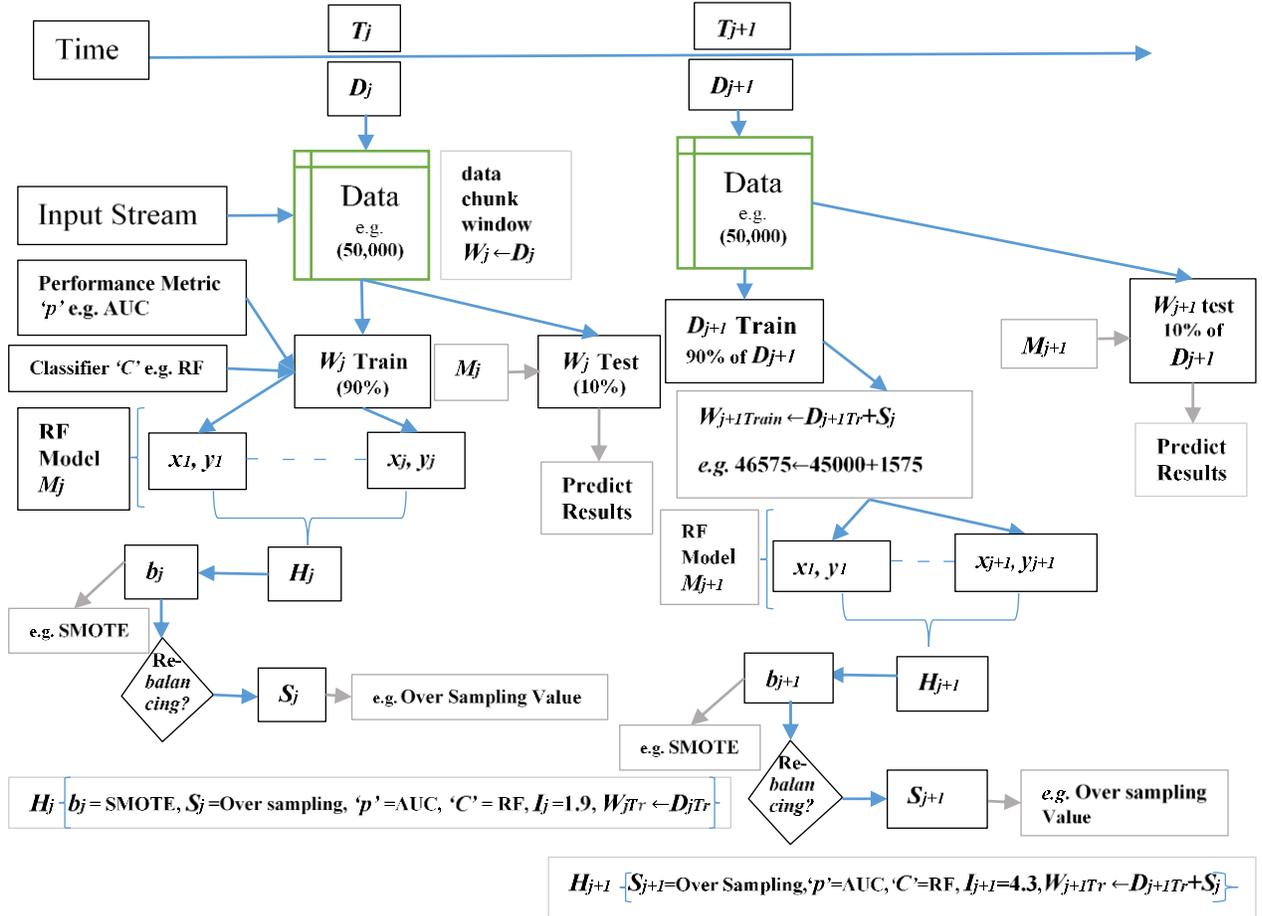

**Fig.1.** Piece-wise incremental data balancing framework for incremental learning.

Ensemble learners in online incremental learning have some interesting properties. Whereby, updating the fixed size base classifier with the arrived new data stream has achieved a good predictive solution [26]. Similarly, the proposed technique of "Piece-Wise Incremental Data Balancing" would increment the size of re-balanced data of the existing window to the arrived new data stream and has produced better prediction results using RF as compared to Batch data.

Our study is unique to implement PWIDB framework and a technique for the incremental data streams. In online learning, large Batch sizes are computationally prohibitive [16]. The technique of PWIDB methodology will train the classifier with incremental re-balancing and will generate efficient models for prediction problems. Our idea is to implement both under-sampling and over-sampling techniques to run parallel on our datasets to adjust a re-balancing strategy dynamically [19] [25], and to ease the incremental learning. For this reason, we incorporated RA [5] in our methodology to implement the automated balancing strategy to resolve the class imbalance problem. This study also examined the RA configuration problems, FD experimental issues such as imbalance ratio of data, size of data [2], and the performance metrics.

IV. EXPERIMENTS

Motivated to implement automated balancing strategy with adapted incremental stream learning, we performed extensive experiments. The steps are detailed and discussed in the following sections.

*A. Datasets*

In this paper, we conducted experiments on real-world highly imbalanced credit card dataset to explore the RA strategy to select the re-balancing technique relevant to the imbalanced dataset for credit card fraud detection. The European Credit Card (ECC) dataset [31] consists of anonymized 284,807 transactions with an Imbalance Ratio (IR) of 1:578, and 0.17% or 492 of fraudulent transactions. The total number of features is 30, and a target feature 'Class' represented with the value 1 in case of fraud and 0 for the legitimate or non-fraud transaction. For this ECC dataset, we used all of the input features and left the optimal features selection job to the classifier for its model generation.

*B. Dealing with imbalanced data*

We have explored eight different balancing techniques to tackle the problem of class imbalance. Random Under Sampling (RUS) [32], Random Over Sampling (ROS) [32], Synthetic Minority Over-Sampling Technique (SMOTE) [33], Tomek link [34], Condensed Nearest Neighbor (CNN) [35], One side Selection (OSS) [36], Edited Nearest Neighbor (ENN) [37] and Neighborhood Cleaning Rule (NCL) [38] techniques. In our work, we implemented an auto-balancing strategy using the Racing Algorithm [5] to run all the above-listed balancing techniques in parallel and selected the best technique based on statistical testing.

*C. Experimental setup*

In this section, we discuss the experimental setup. We ran experiments using R 3.4.1 and RStudio 1.0.153 software. For the experimental setup, we analysed the technique for optimizing the parameters for the Racing Algorithm relevant to the dataset, classification algorithm, and the performance measure.

*D. Racing Algorithm technique*

The Racing Algorithm implemented as an unbalanced package in 'R' enables to terminate prematurely, a configuration if it proves worse than another one. The strategy in the RA is to adapt to the given dataset, performance metric and the classification algorithm [5], along with optimal hyperparameters to select the best balancing method. The challenge of using RA is that the researchers need to run experiments with different hyperparameter options to derive the optimal list of configurations and to allow the RA to select to the best balancing method. In implementing the automated data balancing technique, the ubRacing [5] technique is available in the unbalanced package in R software. It is not easy to implement in other software such as Python, as there is no python toolbox allowing such processing [39]. We analyse the key properties of RA to decide which balancing strategy is statistically significant to use for the streaming datasets to maximize prediction accuracy.

*E. Automated balancing strategy methodology and racing algorithm implementation*

The first set of experiments in Subsection IV-E-1 were conducted to implement automated balancing strategy in Batch mode, which is the method mostly used to deal with imbalance data. In addition, for comparison, we conducted a separate study in a Batch mode without re-balancing strategy for Random Forest classification.

*1) Automated balancing strategy – Batch mode*

In the Batch mode implementation, we ran a few controlled simulated experiments on ECC dataset using RA and RF classification algorithm. For the RF classification algorithm, using a forest of 100 trees is a conventional hyper-parameter number for various applications [40]. In addition, as the number of trees increases the RF stabilize at about 200 trees [41] and the error rates saturate and stabilizes [40]. To resolve the optimization problems for the algorithm design choices, and algorithmic specific parameter is tuning to derive the best performance need to be involved [42]. One common solution is to change one hyper-parameter at a time and to measure its effect on the model's performance [43]. In resolving the algorithm configuration problems [42], we implemented a controlled experimented technique. For the first set of experiments, we kept all parameters constant except the number of trees. Later, we used "max number of experiments" as a variable parameter and kept other parameters constant. We repeated the experiments in similar fashion until we used all the parameters. We even tried combining two parameters as a variable (for e.g., number of trees and maximum number of the experiment), and other parameters as constant. In tables I and II, we listed the best results of the experiments each for AUC and F1 metrics. Similarly, we depicted the experimental results in Fig. 2 and 3.

In addition, we ran experiments using RF classification algorithm without re-balancing the dataset, while selecting similar hyper-parameters from earlier Racing Algorithm experiments. We listed the RF experiments results in the last column of tables I and II each for AUC and F1 metrics.

TABLE I. AUC RESULT FOR ECC DATASET – BATCH LEARNING (AUTOMATED BALANCING STRATEGY – RACING ALGORITHM VS NON RE-BALANCING RANDOM FOREST)

| Exp # | Block Size | Perc Over | Perc Under | Number of trees | Controlled experiment (variable) | Re-Balancing technique | Minority | Majority | IR | Racing Algorithm AUC | Random Forest AUC |
|---|---|---|---|---|---|---|---|---|---|---|---|
| 1 | 284807 | 275 | 275 | 100 | no. of trees | SMOTE | 1296 | 2376 | 1.8 | 0.9917 | 0.9740 |
| 2 | 284807 | 275 | 275 | 75 | no. of trees | SMOTE | 1296 | 2376 | 1.8 | 0.9916 | 0.9659 |
| 3 | 284807 | 275 | 275 | 60 | no. of trees | SMOTE | 1296 | 2376 | 1.8 | 0.9924 | 0.9660 |
| 4 | 284807 | 275 | 275 | 50 | no. of trees | SMOTE | 1296 | 2376 | 1.8 | 0.9921 | 0.9661 |
| 5 | 284807 | 275 | 275 | 25 | no. of trees | SMOTE | 1296 | 2376 | 1.8 | 0.9919 | 0.9579 |
| Prediction Average | | | | | | | | | | 0.9917 | 0.9660 |

TABLE II. F1 RESULT FOR ECC DATASET – BATCH LEARNING (AUTOMATED BALANCING STRATEGY – RACING ALGORITHM VS NON RE-BALANCING RANDOM FOREST)

| Exp # | Block Size | Perc Over | Perc Under | K | Number of trees | Controlled experiment (variable) | Rebalancing tech. | Minority | Majority | IR | Racing Algorithm F1 | Random Forest F1 |
|---|---|---|---|---|---|---|---|---|---|---|---|---|
| 1 | 284807 | 300 | 300 | 5 | 1000 | no. of trees | ROS | 2160 | 255894 | 119 | 0.8908 | 0.8889 |
| 2 | 284807 | 300 | 300 | 5 | 900 | no. of trees | ROS | 2160 | 255894 | 119 | 0.8908 | 0.8889 |
| 3 | 284807 | 300 | 300 | 7 | 800 | no. of trees & K value | ROS | 3024 | 255894 | 85 | 0.8908 | 0.8889 |
| 4 | 284807 | 300 | 300 | 7 | 700 | no. of trees & K value | ROS | 3024 | 255894 | 85 | 0.8908 | 0.8889 |
| 5 | 284807 | 300 | 300 | 7 | 75 | no. of trees & K value | ROS | 3024 | 255894 | 85 | 0.8814 | 0.8793 |
| 6 | 284807 | 300 | 300 | 7 | 50 | no. of trees & K value | ROS | 3024 | 255894 | 85 | 0.8814 | 0.8696 |
| Prediction Average | | | | | | | | | | | 0.888 | 0.884 |

*a) Discussions*

The experimental results as depicted in Fig. 2 reveal that the RA when working with a massive dataset has generated good AUC value of 0.99 with the SMOTE balancing method while RF without re-balancing technique, could generate AUC value of 0.97. Moreover, with re-balancing, the AUC values stabilized at 0.992, the reason being when the training dataset becomes massive the model's accuracy will stop increasing (much) [44]. However, with RF without re-balancing technique the predictions could not improve beyond 0.97 of AUC values. Similarly, the results as depicted in Fig. 3 reveal that using RA for re-balancing technique achieved comparable F1 metric value of 0.891 as compared to RF without re-balancing technique. Moreover, as highlighted with AUC results, F1 values also stabilized around 0.89, and the metric values did not improve even after increasing the number of trees and other parameter values. These experiment results show that automated balancing strategy using RA has improved the prediction results as compared to RF without re-balancing. In addition, the results are convincing that RA works well in Batch mode, which agree to those reported in the literature [5].

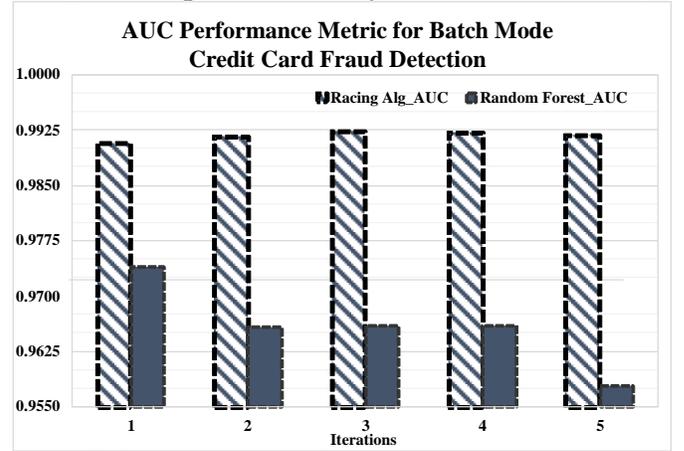
Fig.2. AUC Result for Batch mode Learning.

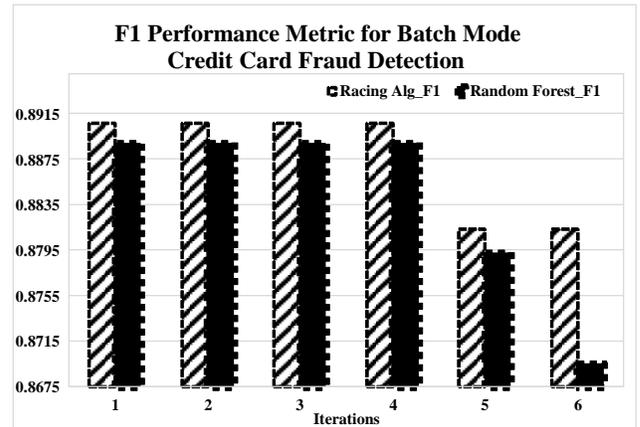
Fig.3. F1 Result for Batch mode Learning.

*2) Accumulative learning technique – streaming*

We conducted the next set of experiments to simulate streaming data to make future predictions. This implies that as soon as the accumulative streaming data builds up, the technique will work for a new hypothesis and for a new classification model generation.

The automated technique to select an appropriate re-balancing technique for data stream learning is a major part of the proposed PWIDB framework. The initial experimental results as discussed in section IV-E-1 have shown that the RA works with any size of data [2]. Now to investigate whether RA can be used for data streaming environment as the automated technique, we have made the historical data accessible to the RA in the form of data streaming windows to assimilate incremental learning to adapt and learn from historical data. We also intend to study the impact of IR on RA with streaming data. Moreover, we also investigated how we can improve automated re-balancing strategy for incremental learning. To validate the significance of accumulative learning technique, we distributed the entire dataset into eight windows and streamed the data incrementally. Whereby each window of data will be distributed as 90% for training and validating the RF model, and the rest 10 % data used for the blind test. In this technique, the data accumulates from the existing window to the subsequent window. This data accumulation technique is applied both for training and test data separately.

For the streaming data learning, we conducted the first set of experiments using the RF algorithm without applying re-balancing techniques to investigate whether RF can also work with streaming data environment. Later, we conducted the second set of experiments by using RA to re-balance the data with accumulative streaming data technique.

TABLE III. RESULT FOR RANDOM FOREST – NON RE-BALANCING (ACCUMULATIVE LEARNING TECHNIQUE)

| Window | Size | Minority | Majority | IR | Minority | Majority | F1 | AUC |
|---|---|---|---|---|---|---|---|---|
| | | | Train & Validation (90%) | | | Test (10%) | | |
| 1 | 50000 | 137 | 44863 | 327.5 | 11 | 4989 | 0.9000 | 1.0000 |
| 2 | 100000 | 205 | 89795 | 438.0 | 18 | 9982 | 0.9412 | 0.9690 |
| 3 | 140000 | 245 | 125755 | 513.3 | 19 | 13981 | 0.9444 | 0.9705 |
| 4 | 180000 | 337 | 161663 | 479.7 | 27 | 17973 | 0.9231 | 0.9788 |
| 5 | 200000 | 357 | 179643 | 503.2 | 28 | 19972 | 0.9057 | 0.9600 |
| 6 | 220000 | 376 | 197624 | 525.6 | 30 | 21970 | 0.9123 | 0.9625 |
| 7 | 250000 | 424 | 224576 | 529.7 | 34 | 24966 | 0.8889 | 0.9812 |
| 8 | 284807 | 455 | 255871 | 562.4 | 37 | 28444 | 0.8824 | 0.9837 |

*a) Discussions*

The experimental results reveal that the accumulative learning technique using RF has improved the performance of the models as compared to Batch data learning. The results as evidenced in Table III has shown improvement over Batch learning as discussed in section IV-E-1-a. Moreover, the prediction average as shown in Table VIII reveals AUC of 0.9757 and F1 of 0.9122 as compared to average Batch data prediction AUC of 0.9660 and F1 of 0.8841 (from Table I and II). The hypothesis here is, as we stream and accumulate the previous Batch data the classifier improved the learning and the model predictions were superior.

Since accumulative learning technique using RF has improved the performance of the models as compared to Batch learning, we intend to apply data re-balancing technique using RA and study its effect on predictive performance. The experimental results for AUC metric are shown in Table IV, and for F1 metric are shown in Table V. The Table VIII shows the average prediction values, whereby RA has AUC prediction of 0.9755 and F1 of 0.9165, whereas RF algorithm without re-balancing has average AUC of 0.9757 and F1 of 0.9122. To summarize, the results using either of the RA implementation are comparable with RF without re-balanced data techniques in streaming data are comparable. Moreover, the RA implementation in streaming data has better AUC prediction value of 0.9755 as compared to 0.9646 [5] for batch data implementation.

TABLE IV. AUC RESULT FOR RACING ALGORITHM (ACCUMULATIVE LEARNING TECHNIQUE FOR RANDOM FOREST)

| Window | Size | Minority | Majority | IR | Re-Balancing tech. | Minority | Majority | IR | Minority | Majority | AUC |
|---|---|---|---|---|---|---|---|---|---|---|---|
| | | | Train & Validation (90%) | | | | | | Test (10%) | | |
| 1 | 45000 | 137 | 44863 | 327.5 | SMOTE | 548 | 1027 | 1.9 | 11 | 4989 | 0.9998 |
| 2 | 90000 | 205 | 89795 | 438.0 | SMOTE | 1435 | 6150 | 4.3 | 18 | 9982 | 0.9678 |
| 3 | 126000 | 245 | 125755 | 513.3 | SMOTE | 1715 | 7350 | 4.3 | 19 | 13981 | 0.9777 |
| 4 | 162000 | 337 | 161663 | 479.7 | SMOTE | 2359 | 10110 | 4.3 | 27 | 17973 | 0.9684 |
| 5 | 180000 | 357 | 179643 | 503.2 | SMOTE | 2499 | 10710 | 4.3 | 28 | 19972 | 0.9662 |
| 6 | 198000 | 376 | 197624 | 525.6 | SMOTE | 1128 | 1504 | 1.3 | 30 | 21970 | 0.9686 |
| 7 | 225000 | 424 | 224576 | 529.7 | SMOTE | 1272 | 1696 | 1.3 | 34 | 24966 | 0.9754 |
| 8 | 256326 | 455 | 255871 | 562.4 | SMOTE | 1365 | 1820 | 1.3 | 37 | 28444 | 0.9798 |

TABLE V. F1 RESULT FOR RACING ALGORITHM (ACCUMULATIVE LEARNING TECHNIQUE FOR RANDOM FOREST)

| Window | Size | Minority | Majority | IR | Re-Balancing tech. | Minority | Majority | IR | Minority | Majority | F1 |
|---|---|---|---|---|---|---|---|---|---|---|---|
| | | | Train & Validation (90%) | | | | | | Test (10%) | | |
| 1 | 45000 | 137 | 44863 | 327.5 | ROS | 274 | 44863 | 163.7 | 11 | 4989 | 0.9000 |
| 2 | 90000 | 205 | 89795 | 438.0 | ROS | 410 | 89795 | 219.0 | 18 | 9982 | 0.9412 |
| 3 | 126000 | 245 | 125755 | 513.3 | ROS | 490 | 125755 | 256.6 | 19 | 13981 | 0.9444 |
| 4 | 162000 | 337 | 161663 | 479.7 | ROS | 674 | 161663 | 239.9 | 27 | 17973 | 0.9231 |
| 5 | 180000 | 357 | 179643 | 503.2 | ROS | 714 | 179643 | 251.6 | 28 | 19972 | 0.9057 |
| 6 | 198000 | 376 | 197624 | 525.6 | ROS | 752 | 197624 | 262.8 | 30 | 21970 | 0.9123 |
| 7 | 225000 | 424 | 224576 | 529.7 | ROS | 848 | 224576 | 264.8 | 34 | 24966 | 0.9063 |
| 8 | 256326 | 455 | 255871 | 562.4 | CNN | 455 | 255594 | 561.8 | 37 | 28444 | 0.8824 |

Hence, accumulative learning techniques have improved the performance of classifiers in the streaming data as compared to Batch data. Moreover, the experimental results with the accumulative learning techniques in streaming data using RA implementation are comparable with RF without re-balanced data; we intend to implement our proposed framework and investigate the PWIDB technique's performance using RA and RF for streaming data.

*3) Piece-wise incremental data balancing technique*

As discussed in the problem formulation section, the PWIDB framework automatically re-balance the current active window and increments the re-balanced data to the next window. To validate our proposal, we conducted experiments both for AUC and F1 metrics, and the results are discussed in the following section.

*a) Discussions*

The experimental results as shown in Tables VI, VII and VIII reveal that PWIDB technique has enhanced the performance of the models and improved the average prediction results. The experimental results as depicted in Table VI and Fig. 4, indicate that incremental re-balancing has improved AUC values and maintained average prediction value of 0.98 (as tabled in VIII) for all streaming windows. With PWIDB technique as learning improves the technique derives new hypothesis $H_t$ at *Time t*. For example, at window *W5* the technique derives a new Hypothesis $H_t$ and learn that it can predict even without any further re-balancing. However, when the technique increments the re-balanced data to the window *W6*, it iterates and suggests that re-balancing the data in the current window has better prediction results. By following this hypothesis, at window *W6* the technique achieved a better AUC measure value of 0.984 for the IR of 2.2. These experimental results for AUC metric reveal that the PWIDB maintained stable average prediction rates as compared to accumulative learning techniques as discussed in section IV-E-2-a. Similarly, the experimental results for F1 performance metric as shown in Tables VII and VIII and Fig. 5 reveal that the PWIDB maintained stable average predictions as compared to accumulative learning techniques as discussed in section IV-E-2-a.

From our analyses, the experimental results with the incremental re-balancing technique implementing RA technique are comparable with RF without re-balancing accumulative learning technique. Applying this new PWIDB technique on highly imbalanced ECC dataset, we discovered comparable AUC performance metric results of RF classification algorithm when compared with non re-balancing RF accumulative learning technique.

Since incremental online learning has gained significant attention in the context of big data [14], the developed incremental data re-balancing technique, using the proposed Piece-Wise Incremental Data Balancing (PWIDB) framework for credit card fraud detection, could handle particularly with highly imbalanced massive datasets.

TABLE VI. AUC RESULT FOR RACING ALGORITHM (INCREMENTAL RE-BALANCING TECHNIQUE FOR RANDOM FOREST)

| Window | Train & Validation (90%) | | | | | | | Test (10%) | | |
|---|---|---|---|---|---|---|---|---|---|---|
| | Size | Minority | Majority | IR | Re-Balancing tech. | Minority | Majority | IR | Minority | Majority | AUC |
| 1 | 45000 | 137 | 44863 | 327.5 | SMOTE | 274 | 44863 | 164 | 11 | 4989 | 0.9998 |
| 2 | 46575 | 616 | 45959 | 74.6 | SMOTE | 4312 | 18480 | 4.3 | 18 | 9982 | 0.9816 |
| 3 | 58792 | 4352 | 54440 | 12.5 | SMOTE | 30464 | 130560 | 4.3 | 19 | 13981 | 0.9775 |
| 4 | 197024 | 30556 | 166468 | 5.5 | SMOTE | 91668 | 183336 | 2.0 | 27 | 17973 | 0.9829 |
| 5 | 293004 | 91688 | 201316 | 2.2 | unbal | 91688 | 201316 | 2.2 | 28 | 19972 | 0.9736 |
| 6 | 311004 | 91707 | 219297 | 2.4 | NCL | 91707 | 219047 | 2.4 | 30 | 21970 | 0.9842 |
| 7 | 337909 | 91755 | 246154 | 2.7 | NCL | 91755 | 245871 | 2.7 | 34 | 24966 | 0.9788 |
| 8 | 368952 | 91786 | 277166 | 3.0 | SMOTE | 275358 | 367144 | 1.3 | 37 | 28444 | 0.9796 |

TABLE VII. F1 RESULT FOR RACING ALGORITHM (INCREMENTAL RE-BALANCING TECHNIQUE FOR RANDOM FOREST)

| Window | Train & Validation (90%) | | | | | | | Test (10%) | | |
|---|---|---|---|---|---|---|---|---|---|---|
| | Size | Minority | Majority | IR | Rebalancing tech. | Minority | Majority | IR | Minority | Majority | F1 |
| 1 | 45000 | 137 | 44863 | 327.5 | ROS | 274 | 44863 | 163.7 | 11 | 4989 | 0.9000 |
| 2 | 90137 | 342 | 89795 | 262.6 | ENN | 342 | 89784 | 262.5 | 18 | 9982 | 0.9418 |
| 3 | 126126 | 382 | 125744 | 329.2 | CNN | 382 | 125344 | 328.1 | 19 | 13981 | 0.9444 |
| 4 | 161726 | 474 | 161252 | 340.2 | unbal | 474 | 161252 | 340.2 | 27 | 17973 | 0.9231 |
| 5 | 179726 | 494 | 179232 | 362.8 | ENN | 494 | 179230 | 362.8 | 28 | 19972 | 0.9057 |
| 6 | 197724 | 513 | 197211 | 384.4 | ROS | 1026 | 197211 | 192.2 | 30 | 21970 | 0.9123 |
| 7 | 225237 | 1074 | 224163 | 208.7 | NCL | 1074 | 223922 | 208.5 | 34 | 24966 | 0.9063 |
| 8 | 256322 | 1105 | 255217 | 231.0 | unbal | 1105 | 255217 | 231.0 | 37 | 28444 | 0.8986 |

TABLE VIII. PREDICTION RESULTS COMPARISON- ECC DATASET

| Window | AUC | | | F1 | | |
|---|---|---|---|---|---|---|
| | Racing Alg_Acc | Racing Alg_Inc_re-balancing | Random Forest_Acc | Racing Alg_Acc | Racing Alg_Inc_re-balancing | Random Forest_Acc |
| 1 | 0.9998 | 0.9998 | **1.0000** | 0.9000 | 0.9000 | 0.9000 |
| 2 | 0.9678 | **0.9816** | 0.9690 | 0.9412 | **0.9418** | 0.9412 |
| 3 | **0.9777** | 0.9775 | 0.9705 | 0.9444 | 0.9444 | 0.9444 |
| 4 | 0.9684 | **0.9829** | 0.9788 | 0.9231 | 0.9231 | 0.9231 |
| 5 | 0.9662 | **0.9736** | 0.9600 | 0.9057 | 0.9057 | 0.9057 |
| 6 | 0.9686 | **0.9842** | 0.9625 | 0.9123 | 0.9123 | 0.9123 |
| 7 | 0.9754 | 0.9788 | **0.9812** | 0.9063 | 0.9063 | 0.8889 |
| 8 | 0.9798 | 0.9796 | **0.9837** | 0.8824 | **0.8986** | 0.8824 |
| Prediction Average | 0.9755 | **0.9822** | 0.9757 | 0.9144 | **0.9165** | 0.9122 |

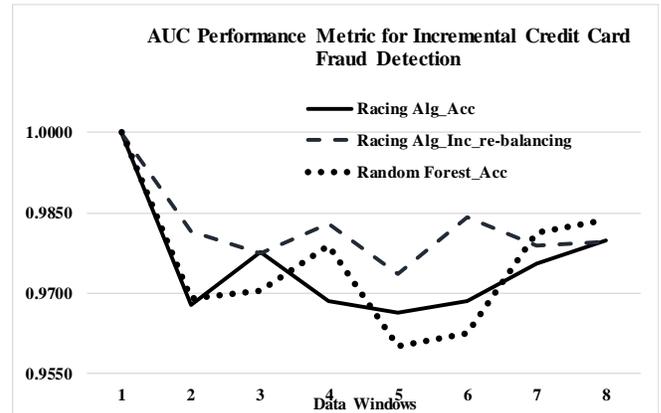

Fig.4. AUC Result for Incremental Learning.

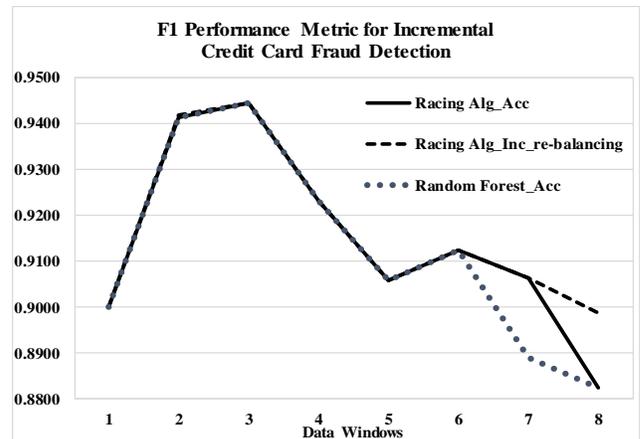

Fig.5. F1 Result for Incremental Learning.

## V. Conclusion

From the experimental studies of the paper, it can be observed that the RA works well in Batch mode however, has little impact when dealing with the massive data stream. The experiments have further demonstrated that RF has performed equally well when compared to RA using the proposed PWIDB framework. Finally, the experiments have shown that the proposed PWIDB framework provides better results when compared to Batch mode, and is more scalable to handle online massive data streams. The PWIDB framework that we have implemented in our research study is suitable to handle incremental learning problems, and for the velocity aspect of big data working with highly imbalanced massive datasets.